\documentclass[]{article}
\usepackage[letterpaper]{geometry}
\usepackage{mtsummit2021}
\usepackage{times}
\usepackage{url}
\usepackage{latexsym}
\usepackage{natbib}
\usepackage{layout}

\usepackage{microtype}
\usepackage{graphicx}
\usepackage{multirow,booktabs}
\usepackage{subcaption}
\usepackage{comment}

\begin{document}

\title{ \bf Scrambled Translation Problem: A Problem ofDenoising UNMT}

\author{\name{\bf Tamali Banerjee} \hfill  \addr{tamali@cse.iitb.ac.in}\\
        \addr{Department of Computer Science and Engineering,
IIT Bombay, India.} \\
\AND
        \name{\bf Rudra Murthy V} \hfill \addr{rmurthyv@in.ibm.com }\\
        \addr{IBM Research Lab, India.} \\
\AND
       \name{\bf Pushpak Bhattacharyya} \hfill \addr{pb@cse.iitb.ac.in}\\
        \addr{Department of Computer Science and Engineering,
IIT Bombay, India.}
}

\maketitle
\pagestyle{empty}

\begin{abstract}
  In this paper, we identify an interesting kind of error in the output of Unsupervised Neural Machine Translation (UNMT) systems like \textit{Undreamt}(footnote). We refer to this error type as \textit{Scrambled Translation problem}. We observe that UNMT models which use \textit{word shuffle} noise (as in case of Undreamt) can generate correct words, but fail to stitch them together to form phrases. As a result, words of the translated sentence look \textit{scrambled}, resulting in decreased BLEU. We hypothesise that the reason behind \textit{scrambled translation problem} is 'shuffling noise' which is introduced in every input sentence as a denoising strategy. To test our hypothesis, we experiment by retraining UNMT models with a simple \textit{retraining} strategy. We stop the training of the Denoising UNMT model after a pre-decided number of iterations and resume the training for the remaining iterations- which number is also pre-decided- using original sentence as input without adding any noise. Our proposed solution achieves significant performance improvement UNMT models that train conventionally. We demonstrate these performance gains on four language pairs, \textit{viz.}, English-French, English-German, English-Spanish, Hindi-Punjabi. Our qualitative and quantitative analysis shows that the retraining strategy helps achieve better alignment as observed by attention heatmap and better phrasal translation, leading to statistically significant improvement in BLEU scores.
\end{abstract}

\section{Introduction}

Training a machine translation system using only the monolingual corpora of the two languages was successfully demonstrated by \citep{artetxe2018iclr,lample2017unsupervised}. They train the machine translation system using denoising auto-encoder (DAE) and backtranslation (BT) iteratively. Recently, pre-training of large language models \citep{conneau2019cross,song2019mass, liu2020multilingual} using monolingual corpus is used to initialize the weights of the encoder-decoder models. These encoder-decoder models are later fine-tuned using backtranslated sentences for the task of Unsupervised Neural Machine Translation (UNMT). While we appreciate language model (LM) pre-training to better initialise the models, it is important to understand the shortcomings of earlier approaches. In this paper, we explore in this direction.


We observe that the translation quality of undreamt models \citep{artetxe2018iclr} suffers partially due to wrong positioning of the target words in the translated sentence. For many instances, though the reference sentence and its corresponding generated sentence are formed with almost the same set of words, the sequence of words is different resulting in the sentence being ungrammatical and/or loss of meaning. This results in a difference in syntax and semantic rules. We define such generated sentences as \textbf{scrambled sentences} and the problem as \textbf{scramble translation problem}. Scrambled sentences can be either \textbf{disfluent} or \textbf{fluent-but-inadequate}. Here, if the LM decoder is not learnt well, we observe disfluent translations. If the LM decoder is learnt well, we observe fluent-but-inadequate translations. An example of fluent-but-inadequate translation will be \textit{`leaving better kids for our planet'} instead of \textit{'leaving better planet for our kids'}. Due to this phenomenon, during BLEU computation n-gram matching lessens, for $n>1$. However, this error is absent in translation generated from recent state-of-the-art systems \citep{conneau2019cross,song2019mass, liu2020multilingual}.



We hypothesise, DAE introduces uncertainty to the previous UNMT \citep{lample2017unsupervised,artetxe2018iclr,artetxe2019effective,wu2019extract} models, specifically to the encoders. It has been observed that encoders are sensitive to the exact ordering of the input sequence \citep{michel-neubig-2018-mtnt,murthy2018addressing,ahmad-etal-2019-difficulties}. By performing random word-shuffle in all the source sentences, encoder may lose important information about the sentence composition. The DAE fails to learn informative representation which affects the decoder resulting in wrong translations generated.

If our hypothesis is true, retraining these previous UNMT system models with noise-free sentences as input should resolve the problem for previous systems \citep{artetxe2018iclr,lample2017unsupervised}. Moreover, using this retraining strategy will not benefit recent approaches \citep{conneau2019cross,song2019mass} as they do not shuffle words of input sentence while training with back-translated data.

In this paper, we prove our hypothesis by showing that a simple \textbf{retraining strategy} mitigates the `scrambled translation problem'. We observe consistent improvements in BLEU score and word-alignment over the denoising UNMT approach by \cite{artetxe2018iclr} for four language pairs. We do not wish to beat the state-of-the-art UNMT systems with pre-training, instead, we demonstrate a limitation of previous denoising UNMT \citep{artetxe2018iclr,lample2017unsupervised} systems and prove why it happens.



\section{Related Work}

Neural machine translation (NMT)~\citep{cho2014learning,sutskever2014sequence,bahdanau2015neural} typically needs lot of parallel data to be trained on. However, parallel data is expensive and rare for many language-pairs. To solve this problem, unsupervised approaches to train machine translation \citep{artetxe2018iclr, lample2017unsupervised, yang2018unsupervised} was proposed in the literature which uses only monolingual data to train a translation system. 

\cite{artetxe2018unsupervised} and \cite{lample2017unsupervised} introduced denoising-based U-NMT which utilizes cross-lingual embeddings and trains a RNN-based encoder-decoder model \citep{bahdanau2015neural}. Architecture proposed by \cite{artetxe2018iclr} contains a shared encoder and two language-specific decoders while architecture proposed by \cite{lample2017unsupervised} contains a shared encoder and a shared decoder. In the approach by \cite{lample2017unsupervised}, the training starts with word-by-word translation followed by denoising and backtranslation. Here, noise in the input sentences in the form of shuffling of words and deletion of random words from sentences was performed. 

\cite{conneau2019cross} (XLM) proposed a two-stage approach for training a UNMT system. The pre-training phase involves training of the model on the combined monolingual corpora of the two languages using Masked Language Modelling (MLM) objective \citep{devlin2019bert}. The pre-trained model is later fine-tuned using denoising auto-encoding objective and backtranslated sentences. \cite{song2019mass} proposed a sequence to sequence pre-training strategy. Unlike XLM, the pre-training is performed via MAsked Sequence to Sequence (MASS) objective. Here, random ngrams in the input is masked and the decoder is trained to generate the missing ngrams in the pre-training phase. The pre-trained model is later fine-tuned using backtranslated sentences.

\cite{murthy-etal-2019-addressing} demonstrated that LSTM encoders of the NMT system are sensitive to the word-ordering of the source language. They considered the scenario of zero-shot translation from language $l_3$ to $l_2$. They train a NMT system for $l_1 \rightarrow l_2$ languages and use $l_1$ - $l_3$ languages bilingual embeddings. This enables the trained model to perform zero-shot translation from $l_3 \rightarrow l_2$. However, if the word-order of the languages $l_1$ and $l_3$ are different, the translation quality from $l_1$ - $l_3$ is hampered.

\cite{michel-neubig-2018-mtnt} have also made a similar observation albeit in the monolingual setting. They observe that accuracy of the machine translation system gets adversely affected due to noise in the input sentences. They discuss various sources of noise with one of them being word emission/insertion/repetition or grammatical errors. The lack of robustness to such errors could be attributed to the sequential processing of LSTM or Transformer encoders. As the encoder processes the input as a sequence and generates encoder representation at each time-step, such errors would lead to bad encoder representations resulting in bad translations generated. Similar observations have also been made by \cite{ahmad-etal-2019-difficulties} for cross-lingual transfer of dependency parsing. They observe that self-attention encoder with relative position representations is more robust to word-order divergence and enable better cross-lingual transfer for dependency parsing task compared to RNN encoders.

\section{Baseline Approach}\label{base}
\begin{figure}[ht!]
\centering
\includegraphics[width=0.9\columnwidth]{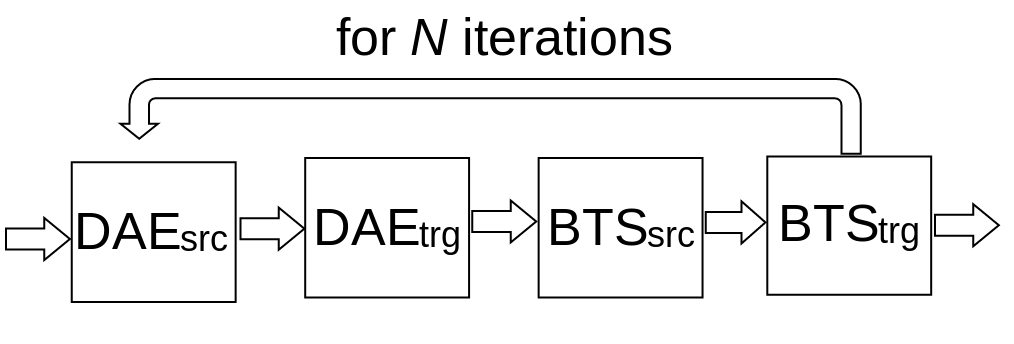}
\caption{Our baseline training procedure: Undreamt. $DAE_{src}$: Denoising of source sentences; $DAE_{trg}$: Denoising of target sentences; $BTS_{src}$: Training with shuffled back-translated source sentences; $BTS_{trg}$: Training with shuffled back-translated target sentences. }
\label{fig:base}
\end{figure}
We use Undreamt \citep{artetxe2018iclr} which is one of the previous UNMT approaches as the baseline for experimentation. \citet{artetxe2018iclr} introduced denoising-based U-NMT which utilize cross-lingual embeddings and train a RNN-based encoder-decoder architecture \cite{bahdanau2015neural}. This architecture contains a shared encoder and two language-specific decoders. Training is a combination of denoising and back translation iteratively as shown in Fig. \ref{fig:base}. By adding noise \citet{artetxe2018iclr} meant shuffling of words of a sentence. Here, shuffling is performed by swapping neighboring words \textit{l/2} times, where \textit{l} is the number of words in the sentence. 4 sub-tasks of the training mechanism are listed below.
(i) $DAE_{src}$: Denoising of source sentences in which we train shared-encoder, source-decoder, and attention with noisy source sentence as input and original source sentence as output.
(ii) $DAE_{trg}$: Denoising of target sentences which trains shared-encoder, target-decoder and attention with noisy target sentence as input and original target sentence as output. 
(iii) $BTS_{src}$: Training shared-encoder, target-decoder, and attention with shuffled back-translated source sentences as input and actual target sentences as output.
(iv) $BTS_{trg}$: Training shared-encoder, source-decoder, and attention with shuffled back-translated target sentences as input and actual source sentences as output.
Here, shuffling is performed by swapping neighboring words \textit{l/2} times, where \textit{l} is the number of words in the sentence.

For completeness, we also experimented with XLM UNMT \citep{conneau2019cross} with initialise the model with MLM objective followed by finetuning it with DAE and BT iteratively. In this approach, they do not add noise with the input sentence while training with backtranslated data.

\section{Proposed Retraining Strategy}
\begin{figure}[ht!]
\centering
\includegraphics[width=0.8\columnwidth]{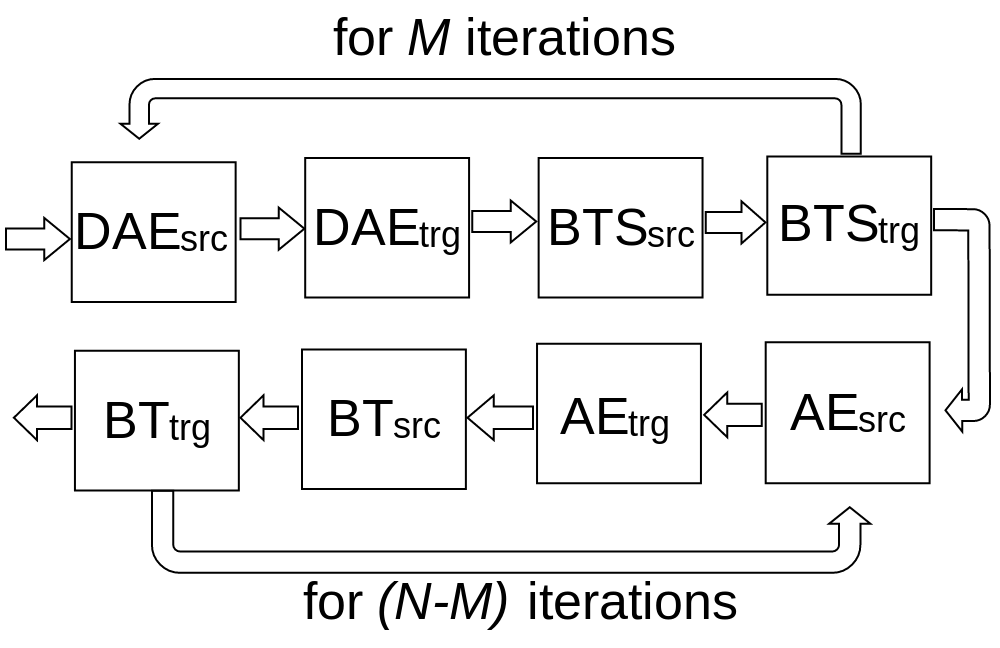}
\caption{ Workflow of Proposed training procedure.  $DAE_{src}$: Denoising of source sentences; $DAE_{trg}$: Denoising of target sentences; $BTS_{src}$: Training with shuffled back-translated source sentences; $BTS_{trg}$: Training with shuffled back-translated target sentences; $AE_{src}$: Autoencoding of source sentences; $AE_{trg}$: Autoencoding of target sentences; $BT_{src}$: Training with shuffled back-translated source sentences; $BT_{trg}$: Training with shuffled back-translated target sentences. 
}
\label{fig:aebt}
\end{figure}

Our proposed strategy to train a denoising-based UNMT system consists of two phases. In the first phase, we proceed with training using denoised sentences similar to the baseline system \citep{artetxe2018iclr} for \textit{M} number of iterations. Adding random shuffling in the input side, however, could introduce uncertainty to the model leading to inconsistent encoder representations. To overcome this, in the second phase, we retrain the model with simple AE and on-the-fly BT using sentences with the correct ordering of words for \textit{(N-M)} iterations as shown in Fig. \ref{fig:aebt}. Here, $N$ is the total number of iterations and $M<N$.  More concretely, this training approach consists of 4 more sub-processes other than the 4 subprocesses of the baseline system. These are: (v) $AE_{src}$: Auto-encoding of source sentences in which we train shared-encoder, source-decoder, and attention. (vi) $AE_{trg}$: Auto-encoding of target sentences in which we train shared-encoder, target-decoder, and attention. (vii) $BT_{src}$: Training shared-encoder, target-decoder, and attention with back-translated source sentences as input and actual target sentences as output. (viii) $BT_{trg}$: Training shared-encoder, source-decoder, and attention with back-translated target sentences as input and actual source sentences as output. The second phase ensures that the encoder learns to generate context representation with information about the correct ordering of words. 
For XLM \citep{conneau2019cross}, we add these 4 subprocesses only with fine-tuning step. We do not change anything in LM pretraining step. 

\section{Experimental Setup}

We test our hypothesis with undreamt as a previous approach and XLM as a SOTA approach. We applied our \textit{retraining strategy} on both the approaches and observed the result.

 For undreamt, we have used monolingual data of six languages, \textit{i.e.} English (en), French (fr), German (de), Spanish (es), Hindi (hi), and Punjabi (pa). Among these languages, Hindi and Punjabi are of SOV word-order where the other four languages are of SVO word order. In our experiments, we choose language-pairs such that the word-order of source language matches with that of target language. We have used the NewsCrawl corpora for en, fr, de of WMT14, and for es of WMT13.  For hi-pa, we use Wikipedia dumps of the august 2019 snapshot for training. The en-fr and en-de models are tested using WMT14 test-data and en-es models using WMT13 test-data, and hi-pa models using ILCI test data \citep{jha2010tdil}.

We have preprocessed the corpus for normalization, tokenization and lowercasing using the scripts available in \textit{Moses}~\citep{koehn2007moses} and \textit{Indic NLP Library}~\citep{kunchukuttan2020indicnlp}, for BPE segmentation using \textit{subword-NMT}~\citep{sennrich2016neural} with number of merge operations set to 50k. 

We use the monolingual corpora to independently train the embeddings for each language using skip-gram model of \textit{word2vec} \citep{mikolov2013distributed}. To map embeddings of two languages to a shared space, we use \textit{Vecmap}\footnote{\url{https://github.com/artetxem/vecmap}} by \citet{artetxe2018robust}. 

We use \textit{undreamt}\footnote{\url{https://github.com/artetxem/undreamt}} tool to train the UNMT system proposed by \citet{artetxe2018iclr}. We train the baseline model untill convergence and noted the number of steps $N$ required to reach convergence. We now train our proposed system for $N/2$ steps and re-train the model after removing denoising noise for the remaining $N/2$ steps. They converge between 500k to 600k steps depending on the language pairs. Further details of dataset and network parameters are available in Appendix.

We also report results on \textit{XLM}\footnote{\url{https://github.com/facebookresearch/XLM}} approach \citep{conneau2019cross}. XLM employs two-stage training of UNMT model. The pre-training stage trains encoder and decoder with masked language modeling objective. The retraining stage employs denoising along with iterative back-translation. However, XLM uses a different denoising (word shuffle) mechanism compared to \citet{artetxe2018iclr}. We replace the denoising mechanism by \citet{conneau2019cross} with the denoising mechanism used by \citet{artetxe2018iclr}. We use the pre-trained models for English-French, English-German, and English-Romanian provided by \citet{conneau2019cross}. We retrain the XLM model until convergence using the denoising approach which makes the baseline system. We later retrain the pre-trained XLM model using our proposed approach where we remove the denoising component after $N/2$ steps.

We report both BLEU scores and n-gram BLEU scores using  \textit{multi-bleu.perl} of Moses. We have tested statistical significance of BLEU improvements \citep{koehnStatTest}. To analyse the systems, we have produced heatmaps of attention generated by the models. 



\begin{table}[!htb]
    \begin{subtable}[b]{.5\linewidth}
    \centering
        \begin{tabular}{c | c c }
    		\toprule
    		\textbf{Language} & \textbf{Baseline } &  \textbf{Retrain} \\
    		\textbf{Pairs} &  \textbf{(Undreamt)} & \textbf{with AE+BT}$\dagger$ \\
    		\midrule
    		 en$\rightarrow$fr & 15.23   & \textbf{17.05}   \\
    		 fr$\rightarrow$en	&  15.99  & \textbf{16.94}  \\
    		 en$\rightarrow$de	& 6.69 &  \textbf{8.03} \\
    		 de$\rightarrow$en & 10.67 &  \textbf{11.66} \\
    		 en$\rightarrow$es	& 15.09 &  \textbf{16.97}   \\
    		 es$\rightarrow$en & 15.33 &  \textbf{17.12}   \\
    		 hi$\rightarrow$pa	& 22.39 &  \textbf{28.61}  \\
    		 pa$\rightarrow$hi &  28.38 & \textbf{33.59}  \\
    		 \bottomrule
        \end{tabular}
        \caption{\label{tab:MainResult} The translation performance using Undreamt-baseline and Undreamt-retraining on en-fr, en-de, en-es, hi-pa test sets (BLEU scores reported). }
        \end{subtable}%
        \hspace{.15in}
        \begin{subtable}[b]{.5\linewidth}
        \begin{tabular}{c | c c }
    		\toprule
    		\textbf{Language} & \textbf{Baseline}  & \textbf{Retrain} \\
    		\textbf{Pairs} &  \textbf{(XLM)}  & \textbf{with AE+BT} \\
    		\midrule
    		 en$\rightarrow$fr & \textbf{33.24} & 31.94 \\
    		 fr$\rightarrow$en & \textbf{31.34}$\dagger$ & 30.79 \\
    		 en$\rightarrow$de & 25.06 & 25.02 \\
    		 de$\rightarrow$en & 30.53 & 30.34  \\
    		 en$\rightarrow$ro & 31.37 & \textbf{31.72}  \\
    		 ro$\rightarrow$en & 29.01 & \textbf{29.96}$\dagger$  \\
    		\bottomrule
    	\end{tabular}
    	\caption{\label{tab:XLMResult} The translation performance using XLM-baseline and XLM-retraining on en-fr, en-de, en-ro test sets (BLEU scores reported).}
	\end{subtable}
	\caption{\label{tab:AllResult} The Translation performance using the Baseline approach and our Approach. Trained for a total of N iterations for all approaches. \textit{Undreamt} and \textit{XLM} results are results from our replication using the code provided by the authors. $\dagger$ indicates statistically significant improvements using paired bootstrap re-sampling \citep{koehnStatTest} for a p-value less than 0.05 .}
\end{table}


\begin{figure*}[!htb]
    \centering
    \begin{subfigure}{0.5\textwidth}
    \includegraphics[width=\columnwidth]{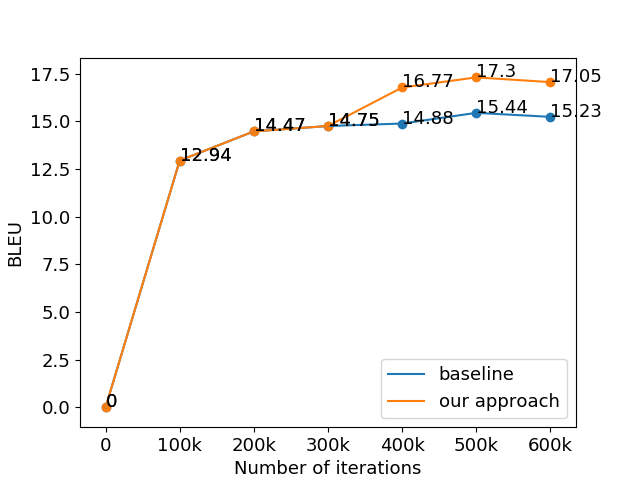}
    \caption{English $\rightarrow$ French}
    \label{fig:enfrbleu}
    \end{subfigure}%
    \begin{subfigure}{0.5\textwidth}
    \includegraphics[width=\columnwidth]{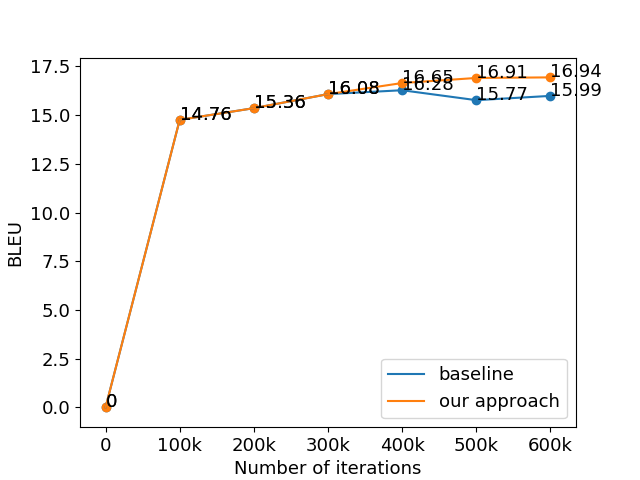}
    \caption{French $\rightarrow$ English}
    \label{fig:frenbleu}
    \end{subfigure}

\caption{Change in translation accuracy using undreamt-baseline vs. our approach with increasing number of iterations for English-French (BLEU scores reported). }
    \label{fig:frenfrbleu}
\end{figure*}
 
 \begin{figure*}[!htb]
    \centering
    \begin{subfigure}{0.5\textwidth}
    \includegraphics[width=\columnwidth]{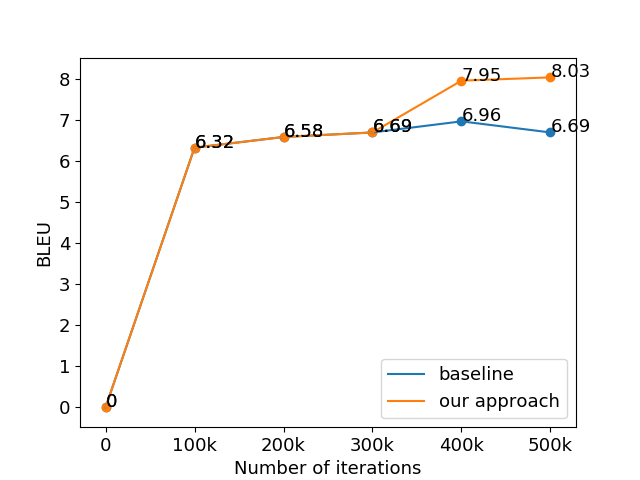}
    \caption{English $\rightarrow$ German}
    \label{fig:endebleu}
    \end{subfigure}%
    \begin{subfigure}{0.5\textwidth}
    \includegraphics[width=\columnwidth]{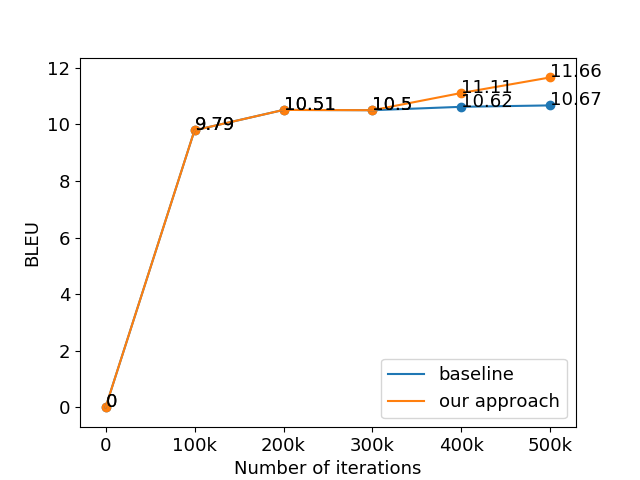}
    \caption{German $\rightarrow$ English}
    \label{fig:deenbleu}
    \end{subfigure}

\caption{Change in translation accuracy using undreamt-baseline vs. our approach with increasing number of iterations for English-German (BLEU scores reported). }
    \label{fig:deendebleu}
\end{figure*}

\begin{figure*}[!htb]
    \centering
    \begin{subfigure}{0.5\textwidth}
    \includegraphics[width=\columnwidth]{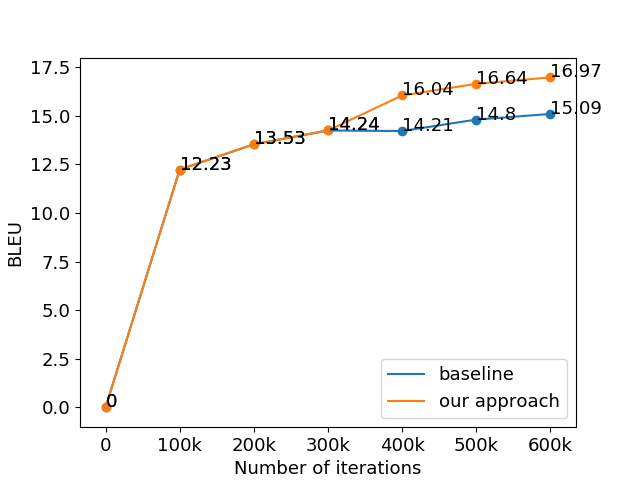}
    \caption{English $\rightarrow$ Spanish}
    \label{fig:enesbleu}
    \end{subfigure}%
    \begin{subfigure}{0.5\textwidth}
    \includegraphics[width=\columnwidth]{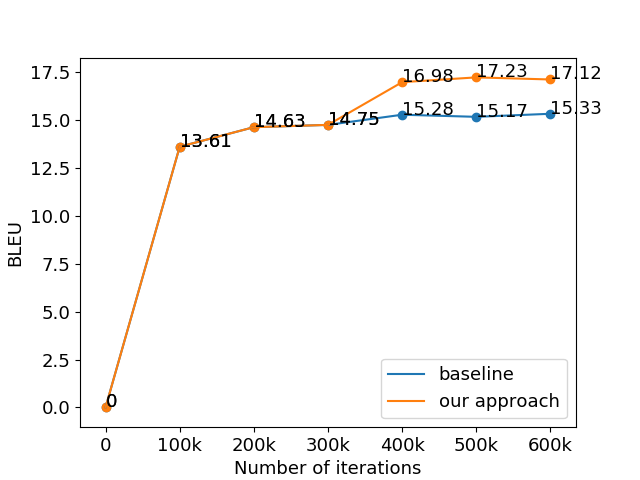}
    \caption{Spanish $\rightarrow$ English}
    \label{fig:esenbleu}
    \end{subfigure}

\caption{Change in translation accuracy using undreamt-baseline vs. our approach with increasing number of iterations for English-Spanish (BLEU scores reported). }
    \label{fig:esenesbleu}
\end{figure*}

\begin{figure*}[!htb]
    \centering
    \begin{subfigure}{0.5\textwidth}
    \includegraphics[width=\columnwidth]{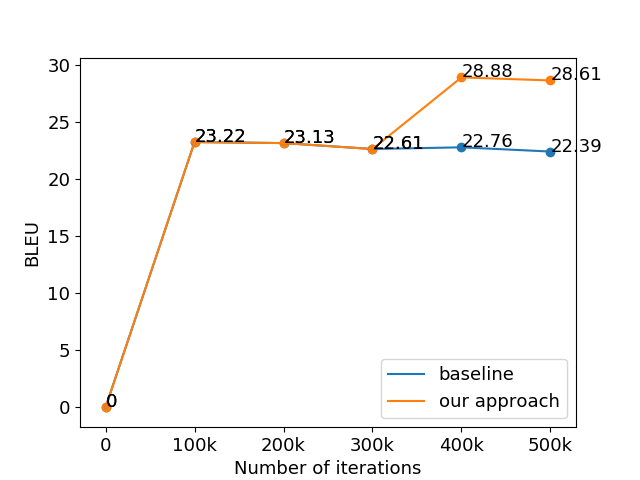}
    \caption{Hindi $\rightarrow$ Punjabi}
    \label{fig:hipableu}
    \end{subfigure}%
    \begin{subfigure}{0.5\textwidth}
    \includegraphics[width=\columnwidth]{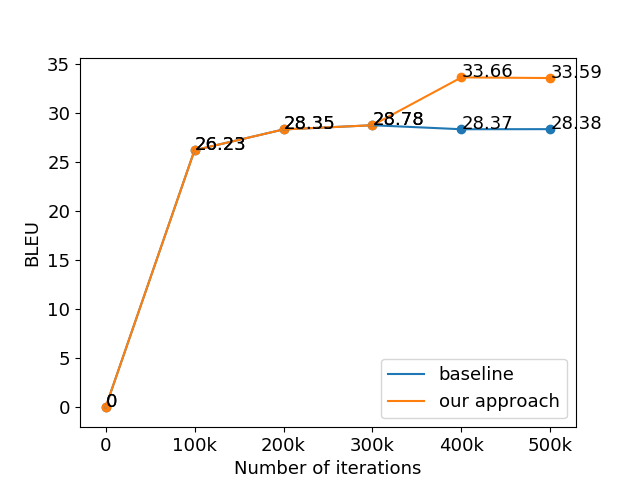}
    \caption{Punjabi $\rightarrow$ Hindi}
    \label{fig:pahibleu}
    \end{subfigure}

\caption{Change in translation accuracy using undreamt-baseline vs. our approach with increasing number of iterations for Hindi-Punjabi (BLEU scores reported). }
    \label{fig:pahipableu}
\end{figure*}

\begin{table}[!htb]
    \centering
    \resizebox{0.7\columnwidth}{!}{
    \begin{tabular}{l | r r r r}
		\toprule
		\textbf{Language}  & \multirow{2}{*}{\textbf{$\Delta$ BLEU-$1$ }} & \multirow{2}{*}{\textbf{$\Delta$ BLEU-$2$ }} & \multirow{2}{*}{\textbf{$\Delta$ BLEU-$3$ }} & \multirow{2}{*}{\textbf{$\Delta$ BLEU-$4$ }} \\
		\textbf{Pairs} &  &  &  & \\
		\midrule
		 en$\rightarrow$fr & 0.00 & 4.50 & 8.85 & 11.67 \\
		 fr$\rightarrow$en & 2.17 & 5.53  & 7.48 & 10.90 \\
		 en$\rightarrow$de & 17.44 & 11.71 & 17.07 & 25.00 \\
		 de$\rightarrow$en & 1.75 & 6.87 & 12.12 & 13.33 \\
		 en$\rightarrow$es & 1.75 & 6.88 & 12.04 & 20 \\
		 es$\rightarrow$en & 3.20 & 9.13 & 14.85 & 21.15 \\
		 hi$\rightarrow$pa & 7.49 & 24.48 & 32.71 & 46.39 \\
		 pa$\rightarrow$hi & 4.30 & 15.89 & 24.12 & 30.56 \\
		\bottomrule
	\end{tabular}
	}
	\caption{\label{tab:n-bleu} Improvements in n-BLEU (represented in \%) on using our approach over baseline for en-fr, en-de, en-es, hi-pa test sets. }
\end{table}

\begin{figure*}[!htb]
    \centering
    \resizebox{\textwidth}{!}{
    \includegraphics{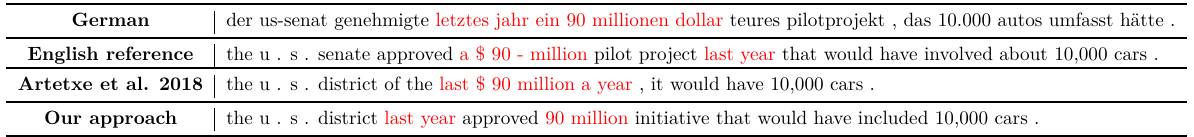}
    }
    \caption{Sample translation of German $\rightarrow$ English translation models.}
    \label{fig:deenexample}
    \end{figure*}

\begin{figure*}[!htb]
    \centering
    \resizebox{\textwidth}{!}{
    \includegraphics{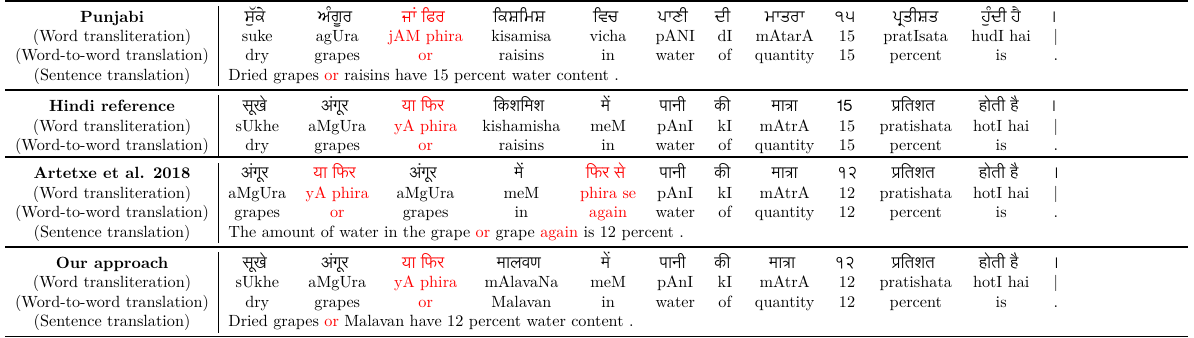}
    }
    \caption{Sample translation of Punjabi$\rightarrow$Hindi translation models.}
    
    \label{fig:pahiexample}
    \end{figure*}   
    
\begin{figure*}[!htb]
    \centering
    \resizebox{\textwidth}{!}{
    \includegraphics{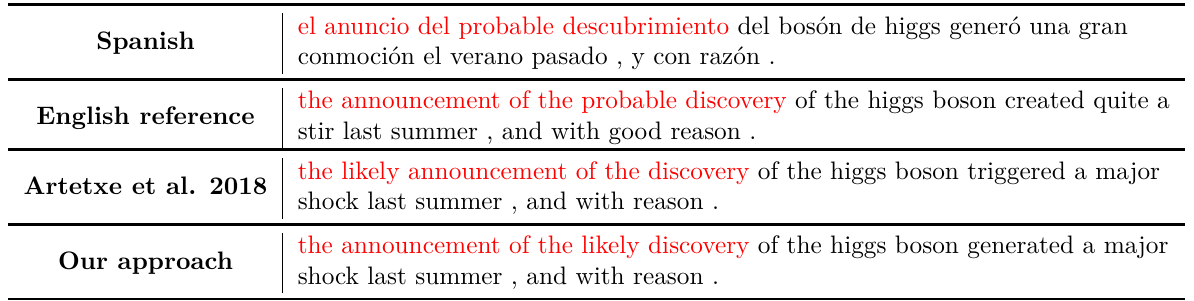}
    }
    \caption{Sample translation of Spanish $\rightarrow$ English translation models.}
    \label{fig:esenexample}
    \end{figure*}

\begin{figure*}[!htb]
    \centering
    \resizebox{\textwidth}{!}{
    \includegraphics{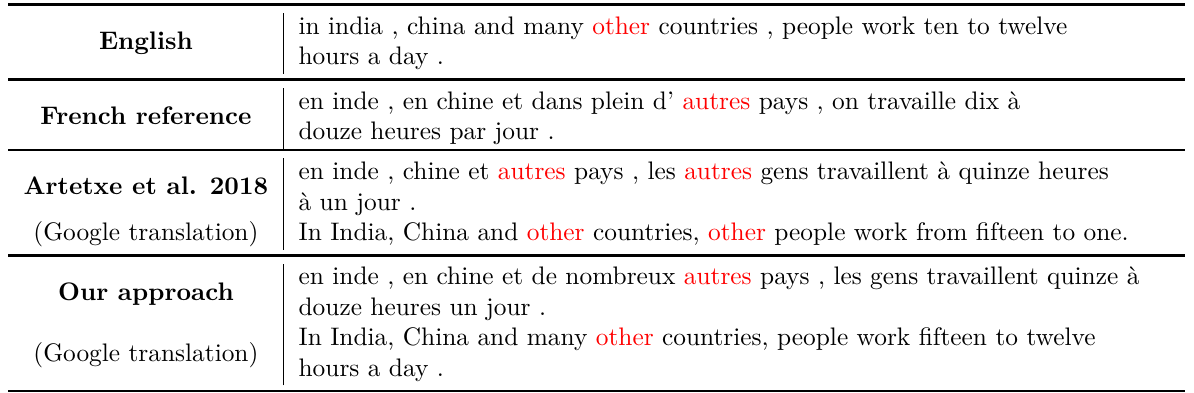}
    }
    \caption{Sample translation of English $\rightarrow$ French translation models.}
    \label{fig:enfrexample}
    \end{figure*}

\begin{figure*}[ht!]
\centering

\begin{subfigure}{0.5\textwidth}
\includegraphics[width=\columnwidth]{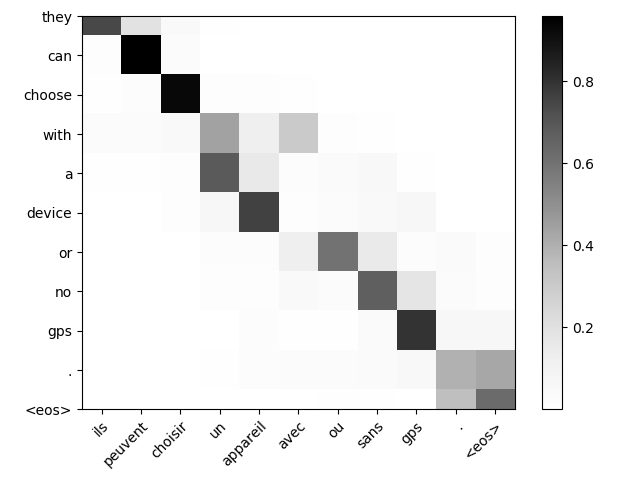}
\caption{Using baseline approach}
\label{heatmap1baseline}
\end{subfigure}%
\begin{subfigure}{0.5\textwidth}
\includegraphics[width=\columnwidth]{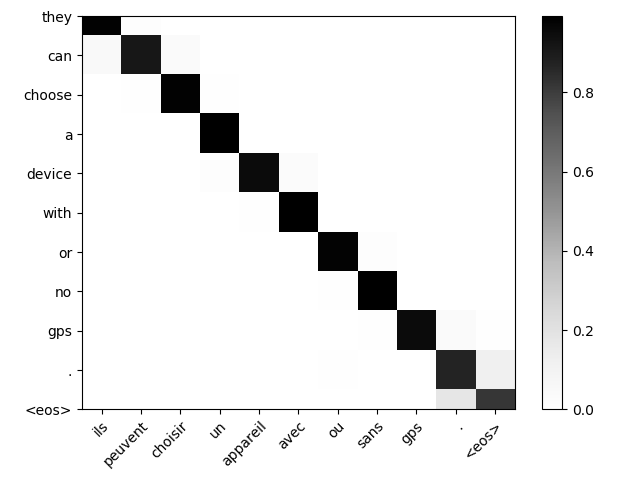}
\caption{Using our approach}
\label{heatmap1unfr}
\end{subfigure}
\caption{Attention heatmaps of a French$\rightarrow$English translation.}
\label{heatmap1}

\end{figure*}

\section{Results and Analysis}
Table \ref{tab:AllResult} reports BLEU score of the trained models using the undreamt  \citep{artetxe2018iclr} and XLM  \citep{conneau2019cross} and retraining them with our approach. \textit{Undreamt} and \textit{XLM} results are results from our replication using the code provided by the authors. In Table \ref{tab:MainResult} we observe that the proposed re-training strategy of AE used in conjunction with BT results in statistically significant improvements (p-value $<$ 0.05) across all language pairs when compared to the undreamt baseline approach \citep{artetxe2018iclr}.

We report results on XLM \citep{conneau2019cross} with our \textit{retraining} approach in Table \ref{tab:XLMResult}. XLM is one of the  state-of-the-art (SOTA) UNMT approaches for these language pairs. The approach by XLM \citep{conneau2019cross} does not add noise to the input backtranslated sentence during training. Therefore, our retraining strategy does not benefit here. We also observe robustness of the pre-trained language models to the scrambled translation problem.


Fig. \ref{fig:frenfrbleu}, \ref{fig:deendebleu}, \ref{fig:esenesbleu} and \ref{fig:pahipableu} show changes in BLEU scores of intermediate UNMT models with increasing number of iterations on test-data. We observe that our proposed approach leads to increase in BLEU score in the re-training phase as the denoising strategy is removed. The baseline system suffers from drop in BLEU score due to denoising strategy introducing ambiguity into the model.


\subsection{Quantitative analysis}
We hypothesize that the baseline UNMT model using DAE is able to generate correct word translation but fails to stitch them together to generate phrases. To validate the hypothesis, we calculate the percentage improvement on using our approach over the baseline system in terms of individual n-gram (n=1,2,3,4) specific BLEU scores for each language-pair and a particular value of \textit{n}. The results presented in Table \ref{tab:n-bleu} indicate that our method achieves higher improvements in n-gram BLEU for higher $n-$grams $(n>1)$ compared to the improvement in n-gram BLEU for lower values of \textit{n}, indicating better phrasal matching. This could be attributed to the proposed approach not suffering from the \textit{scrambled translation problem} introduced by the DAE. 

\subsection{Qualitative analysis} 
We observe several instances where our proposed approach results in better translations compared to the baseline.
On manual analysis of translation outputs generated by the baseline system, we have found out some instances of \textit{scrambled translation problem}. 


Due to uncertainty introduced by shuffling of words before training, the baseline model chooses to generate sentences that are more acceptable by a language model.
Fig \ref{fig:deenexample} shows such an example in our test data. Here, two German phrases \textit{`ein 90 millionen'} (\textit{`a 90 million'}) and \textit{`letztes jahr'} (\textit{`last year'}) are mixed up and translated as \textit{`last \$ 90 million a year'} in English. However, our approach handled the issue correctly. 

 Fig \ref{fig:pahiexample} shows an example of a situation where the baseline model prefers to generate a word in multiple probable positions. Here, the source Punjabi sentence consists of a phrase \textit{`jAM phira'} (\textit{`or'}) meaning \textit{`yA phira'}(\textit{`or'}) in Hindi. In the translation produced by the baseline model, the correct phrase is generated along with the word \textit{`phira'} wrongly occurring again forming another phrase \textit{`phira se'} (\textit{`again'}). Note that, both the phrases are commonly used in Hindi. In Fig \ref{fig:esenexample}, the model trained on baseline system produced the word \textit{`likely'}, which is a synonym of `probably', in the wrong position. In Fig \ref{fig:enfrexample}, the model trained on baseline system produced the word \textit{`autres'}(\textit{`other'}) in the multiple positions.

\paragraph{Attention Analysis:}
Attention distributions generated by our proposed systems have lesser confusion when compared with the attention distribution generated by baseline systems, as shown in Heatmaps of Fig. \ref{heatmap1}. Production of word-aligned attention distribution was easy for the attention models, which we retrained on sentences without noise. 

\section{Conclusion and Future work}
In this paper, we addressed `scrambled translation problem', a shortcoming of previous denoising-based UNMT approaches like \textit{UndreaMT} approach \citep{artetxe2018iclr,lample2017unsupervised}. We demonstrated that adding shuffling noise to all input sentences is the reason behind it. Our simple \textit{retraining} strategy, \textit{i.e.} retraining the trained models by removing the denoising component from auto-encoder objective (AE), results in significant improvements in BLEU scores for four language pairs. We observe larger improvements in n-gram specific BLEU scores for higher value of \textit{n} indicating better phrasal translations. We also observe robustness of the pre-trained language models to the scrambled translation problem. We would also like to explore applicability of our approach in other ordering-sensitive DAE-based tasks.

\bibliographystyle{apalike}
\bibliography{mtsummit2021}

\end{document}